\documentclass[10pt,twocolumn,letterpaper]{article}

\usepackage[pagenumbers]{cvpr} % To force page numbers, e.g. for an arXiv version

\usepackage{graphicx}
\usepackage{amsmath, bm}
\usepackage{amssymb}

\usepackage{caption}
\usepackage{subcaption}
\usepackage[normalem]{ulem}
\usepackage[accsupp]{axessibility}  % Improves PDF readability for those with disabilities.
\newcommand{\naively      }     {{na\"{\i}vely}}

\newcommand{\norm}[1]{\left\lVert#1\right\rVert}

\newcommand{\model}{\textsc{RAG-Gesture}}

\definecolor{RDcolor}{rgb}{0.5, 0.1, 0.8}

\definecolor{bronze}{rgb}{1,1,0.6}
\definecolor{silver}{rgb}{0.969,0.796,0.600}
\definecolor{gold}{rgb}{0.941,0.592,0.600}
\newcommand{\gold}[1]{\colorbox{gold}{{#1}}}
\newcommand{\silver}[1]{\colorbox{silver}{{#1}}}

\definecolor{cvprblue}{rgb}{0.21,0.49,0.74}
\usepackage[pagebackref,breaklinks,colorlinks,allcolors=cvprblue]{hyperref}

\def\confName{CVPR}
\def\confYear{2025}

\title{Retrieving Semantics from the Deep: an RAG 
Solution for Gesture Synthesis} 

\author{M. Hamza Mughal$^{1}$\hspace{1.8em} Rishabh Dabral$^{1}$\hspace{1.8em} 
Merel C.J. Scholman$^3$\hspace{1.8em} \\Vera Demberg$^{1,2}$\hspace{1.8em}
Christian Theobalt$^{1,2}$\\
$^1$Max Planck Institute for Informatics, SIC 
\hspace{1em}
$^2$Saarland University
\hspace{1em}
$^3$Utrecht University\\
\vspace{-10pt}
\\
}
\begin{document}
\twocolumn[{ 
\renewcommand\twocolumn[1][]{#1} 
\maketitle 
\begin{center} 
    \vspace{-22pt} 
    \includegraphics[width=\linewidth]{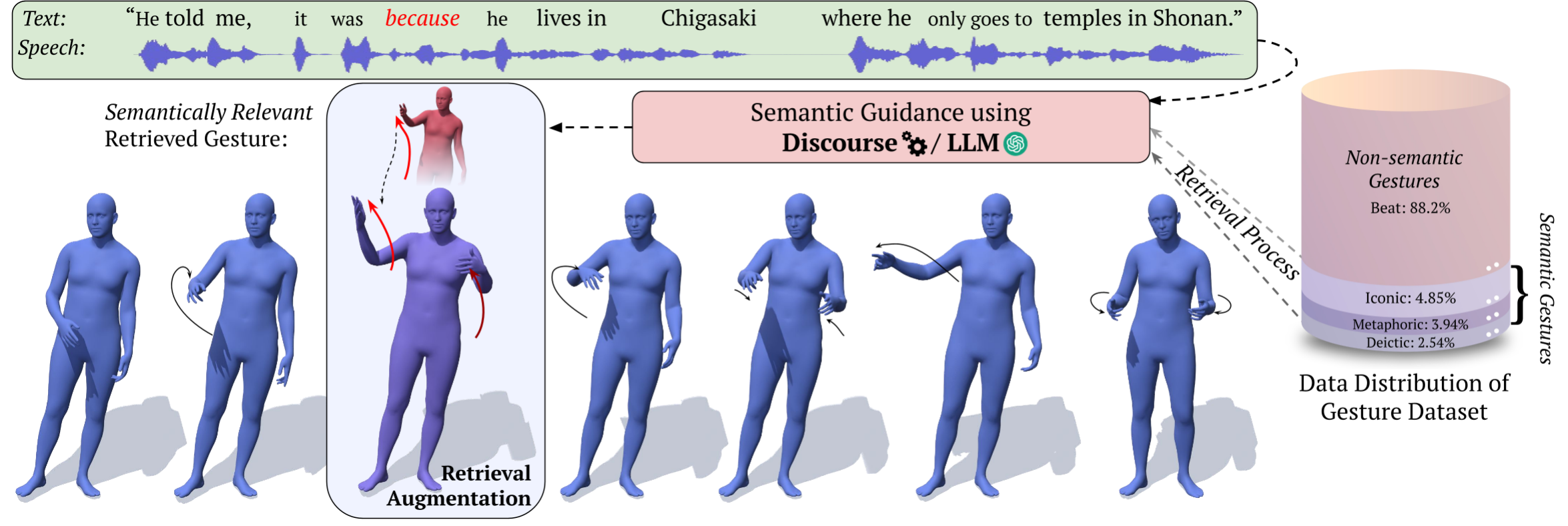}
    \vspace{-22pt} 
    \captionof{figure}{Our \model~approach produces \textbf{semantically meaningful gestures} by leveraging explicit knowledge to retrieve exemplar gestures from the sparse semantic data~\cite{liu2022beat} and \textbf{guiding the diffusion-based generation process} through \textbf{Retrieval Augmentation}. 
    }
    \label{fig:TEASER}
\end{center} 
}]
\begin{abstract}
% \vspace{-1.2pt}
Non-verbal communication often comprises of semantically rich gestures that help convey the meaning of an utterance.
Producing such semantic co-speech gestures has been a major challenge for the existing neural systems that can generate rhythmic beat gestures, but struggle to produce semantically meaningful gestures. 
Therefore, we present \model, a diffusion-based gesture generation approach that leverages Retrieval Augmented Generation (RAG) to produce natural-looking and semantically rich gestures.
Our neuro-explicit gesture generation approach is designed to produce semantic gestures grounded in interpretable linguistic knowledge.  
We achieve this by using explicit domain knowledge to retrieve exemplar motions from a database of co-speech gestures.
Once retrieved, we then inject these semantic exemplar gestures into our diffusion-based gesture generation pipeline using DDIM inversion and retrieval guidance at the inference time without any need of training.
Further, we propose a control paradigm for guidance, that allows the users to modulate the amount of influence each retrieval insertion has over the generated sequence.
Our comparative evaluations demonstrate the validity of our approach against recent gesture generation approaches.
The reader is urged to explore the results on~\href{https://vcai.mpi-inf.mpg.de/projects/RAG-Gesture/}{our project page}.
\end{abstract}    
\section{Introduction}
\label{sec:intro}
Human communication is a complex and multifaceted process that involves both verbal and non-verbal elements.
Non-verbal communication  comprises \textit{co-speech gestures}, which are defined as body and hand movements that are temporally aligned and semantically integrated with the speech~\cite{mcneill1992handmind}.
Gestures convey information in tandem with speech and language, and can carry additional meaning that enhances the semantic construct of the message.
Generating meaningful full-body gestures is therefore important for the communicative efficacy of virtual humans in telepresence and content-creation domains.
\par
McNeill~\cite{mcneill2005gesturethought, mcneill1992handmind} categorizes gestures into
\textit{beat gestures}, which are rhythmic movements driven by prosody, and \textit{semantics-driven gestures}, which carry the communicative intent.
Semantic gestures are context-driven and bear a specific meaning that complements the utterance~\cite{mcneill1992handmind,HBM_bernard}.
Rule-based approaches retrieve these semantic gestures and combine them for generation, while neural methods leverage the data to learn the synthesis process.
However, the former tends to result in an unnatural outcome and the latter struggles to generate semantic gestures because these gestures, while being a part of the existing datasets, occur rather sporadically compared to the rhythmic beat gestures~\cite{nyatsanga2023comprehensive}. 
\par
In this work, we
%The current work
address this issue by framing the problem as a Retrieval Augmented Generation (RAG) task. 
We propose \model~-- a diffusion-based gesture synthesis approach that generates natural and semantically rich gestures
by retrieving context-appropriate exemplars from a database and injecting them into the gesture generation process.
Motivated by the insights from Neff~\cite{neff2016hand}, we decompose the generation problem into two tasks: \textit{specification}, which ascertains ``what gesture'' to produce for ``which word'' in the speech, and \textit{animation}, which determines ``how to generate'' that specified gesture.
\par
To solve the specification problem, we explicitly retrieve the relevant exemplar motions from a gesture database using either of the two retrieval algorithms proposed in~\cref{subsec:retrieval-algos}.
More specifically, we show that the chain-of-thought reasoning abilities of Large Language Models~\cite{openai2024gpt, zhao2023llmsurvey} can be exploited to extract which parts of the utterance would likely be semantically gestured, and what `type' of semantic gesture (\textit{iconic, deictic, metaphoric, etc.}) the phrase would invoke.
We also demonstrate that one could, alternatively, use specific linguistic elements to base the retrieval on. 
In particular, we focus on discourse connectives such as \textit{because}, \textit{while}, and \textit{on one hand}, which have been shown to affect the gestural patterns~\cite{mcneill2014discourse, calbris2011elements}.
\par
The animation problem, in our context, requires integrating the retrieved gesture motions (corresponding to a subset of the sentence) into the overall motion generated for the full sentence/utterance (see~\cref{fig:TEASER}). 
To this end, we design an inference-time approach to augment the gesture generation capability of diffusion models with retrieved gesture clips.
Our approach allows us to surgically integrate the semantic gestures into the generation process of the base model, %
while not affecting the synthesis for the remaining parts of the utterance.
%\
\par
\begin{figure}
    \centering
    \includegraphics[width=\linewidth]{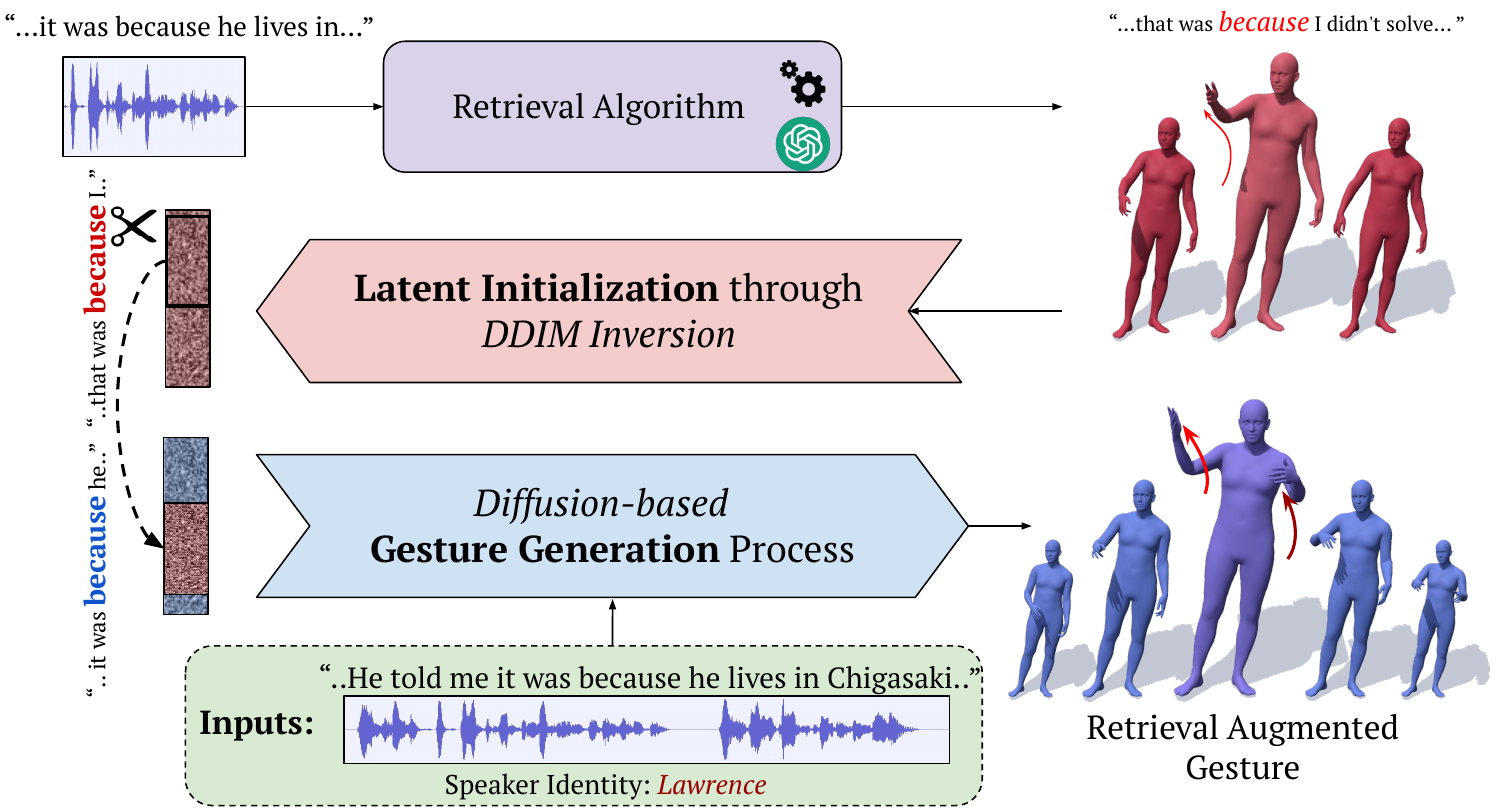}
    \caption{
    \textbf{Overview.} 
    Our approach retrieves example gestures on semantically important words in speech and inserts those examples into the generated gesture by using them to guide the generation.}
    \vspace{-15pt}
    \label{fig:overview}
\end{figure}
Through comprehensive numerical and perceptual evaluations, we demonstrate the effectiveness of our proposed neuro-explicit approach compared with purely neural state-of-the-art approaches like~\cite{liu2024emage, zhang2023remodiffuse, ng2024audio2photoreal, liu2022beat} that typically struggle to synthesize semantically meaningful gestures.
To summarize, our core contributions are as follows:
\begin{itemize}
    \item We introduce \model~-- a diffusion-based approach which leverages RAG to generate natural-looking and semantically rich gestures by injecting meaningful motion exemplars into the gesture generation process.
    \item We introduce multiple retrieval algorithms to extract semantically relevant exemplars from the sparse semantic data distribution by leveraging either the linguistic structure of utterances or LLM-based gesture type prediction.
    \item Further, we propose a novel inference-time RAG approach for our framework and demonstrate how gesture exemplars can be transferred onto the generated motion locally through the combination of Latent Initialization and Retrieval Guidance technique, which provides a way to control the influence of the retrieval insertion.
\end{itemize}

\section{Related Work}
\label{sec:related-works}
We first provide an overview of gesture synthesis approaches, followed by a discussion of relevant approaches from RAG literature.
Lastly, we discuss recurrent patterns of gestures used to express discourse structure, which motivates the design of the retrieval-based method.
\subsection{Co-Speech Gesture Synthesis}
Co-speech gestures consist of hand and arm movements that occur in sync with speech and help convey meaning and structure discourse.
Earlier methods in co-speech gesture generation~\cite{cassell2001rulebeat, thiebaux2008rulesmartbody} perform well on semantics and appropriateness of gestures~\cite{nyatsanga2023comprehensive}.
These systems retrieve the most appropriate gesticulation units for given speech, resulting in high semantic alignment. 
However, since they insert predefined units into animation, the synthesis quality of these systems tends to be unnatural.
To mitigate the lack of naturalness, recent learning-based methods~\cite{kucherenko2019analyzing, alexanderson20style, ferstl2020adversarial, habibie21learning} employ deep networks to convert speech input to gesture output and aim to ensure smooth motions. 
These methods are mostly data-driven and rely on learning speech-to-gesture matching through large scale gesture datasets~\cite{liu2022beat, yoon2019robots, ferstl18investigating, ghorbani2022zeroeggs}.
Most of these recent methods~\cite{yoon2020trimodal, kucherenko2019analyzing, ahuja2020style, yi2023talkshow} suffer from reduced communicative efficacy and low semantic alignment, due to bad generalization to semantic co-speech gestures in the dataset.
Furthermore, generation and reproduction of semantically appropriate gestures corresponding to speech still remains a challenge for these systems \cite{yoon2022genea}.
\par
A few recent approaches directly addressed this problem.
Kucherenko~\etal~\cite{kucherenko2021speech2properties2gestures} use speech to predict the gesture properties, which inform the model on what type of gesture to generate.
Instead, our framework proposes to use RAG to follow the gesture type information instead of taking the properties as conditioning input.
%
% Liang~\etal~\cite{liang2022seeg} decouple semantic irrelevant cues like rhythmic beats and volume to generate plausible gestures.
% 
% 
Mughal~\etal~\cite{mughal2024convofusion} generate semantic gestures by providing word-level semantic control and use it to sample semantically meaningful gestures.
Zhi~\etal~\cite{zhi2023livelyspeaker} initialize diffusion-based generation with text-aligned embeddings by adding noise to them (similar to inpainting~\cite{lugmayr2022diffinpaint}).
In contrast, our approach inverts semantic exemplars into the diffusion latent space and uses them to initiate the denoising process. 
SemanticGesticulator~\cite{zhang2024semanticgesture} is a generative framework which retrieves semantic gestures from a curated motion library and aligns them with rhythmic gestures through a training-based approach.
In comparison, our framework performs inference-time RAG using a diffusion model rather than fine-tuning the gesture generator to follow the retrieval.
Secondly, our approach also utilizes prosody information for every retrieval instead of only using textual transcription.
Thirdly, instead of relying on a curated library, our approach introduces explicit algorithms to retrieve semantics from the existing data~\cite{liu2022beat}.
Lastly, our framework is flexible enough to leverage any retrieval algorithm at inference time.
\subsection{Retrieval Paradigms for Generative Methods}
Employing database retrieval for improved animation performance is a prevalent paradigm in motion and gesture synthesis.
Classical~\cite{buttner2015motion, clavet2016motion} and learning-based motion matching approaches~\cite{holden2018character, holden2020learned} have been applied for better character control in motion synthesis~\cite{starke2024categorical}.
Diffusion-based text-to-motion synthesis frameworks also deal with the problem of semantic control at a global level~\cite{zhang2023remodiffuse} or at the local keyframe level~\cite{goel2024iterative, huang2024controllable}.
Zhang~\etal~\cite{zhang2023remodiffuse} apply RAG to diffusion-based text-to-motion synthesis using global text similarity for retrieval.
Unlike text-to-motion, gesture generation is affected by language content and prosody at varying levels of semantic granularity. 
That means one can neither perform global semantic matching between the retrieved and generated motion~\cite{zhang2023remodiffuse} nor simply copy and paste retrieved motion parts onto the keyframe locations~\cite{goel2024iterative} without considering the differences in context.
\par
Gesture synthesis approaches have also utilized database retrieval both in classical and learning based paradigms.
Early approaches~\cite{neff2008statgesturestyle, kipp2005gesture, bergmann2009gnetic} utilize statistics to retrieve gestures from a database of gesticulation units.
A few learning-based approaches combine rule-based retrieval and deep learning-based synthesis to improve upon the semantic quality of gesture generation.
ExpressGesture~\cite{ferstl2021expressgesture} introduces a database-driven framework to ensure that generated gestures retain the expressive and defined gesture form.
Habibie~\etal~\cite{habibie2022motion} employ nearest neighbor search to retrieve most appropriate gestures and provide a style-control mechanism over the generative framework.
In comparison, our retrieval approach is not limited to style or select keywords but is grounded in explicit rules for semantic retrieval that are driven from linguistic and gestural structure.
\subsection{The effect of Linguistic Dynamics on Gestures}
The goal of our retrieval approach is to extract semantic gestures from the database. Our proposed design is therefore motivated by the relation between language dynamics and co-speech gestures.
Laparle~\cite{laparle2021discourse} demonstrates how gesture features like hand shape or orientation are important to distinguish between topics in discourse.
Gestures can also indicate the relations between different parts of the message.
For example, a contrast relation can be expressed by the lexical marker \textit{on the one hand...on the other hand}, framing the two contrasting arguments as items placed on separate hands \citep{hinnell2019verbal}. 
The raised index finger draws the attention of the interlocutor to new and important topics, and is commonly used to express exceptions or concessions (which can be expressed linguistically using \textit{however})  \citep{bressem2014repertoire,kendon2004gesture,inbar2022raised}.
\par
In sum, gestures can be used to convey the semantic structure of a message, but existing generation works fail to reproduce these context-driven patterns~\cite{yoon2022genea}, which results in repetitive gestures.
We attempt to leverage the linguistic information in our retrieval framework.
This framework can, in turn, improve semantic grounding of generated gestures through RAG based insertion.
\begin{figure*}
    \centering
    \includegraphics[width=0.95\linewidth]{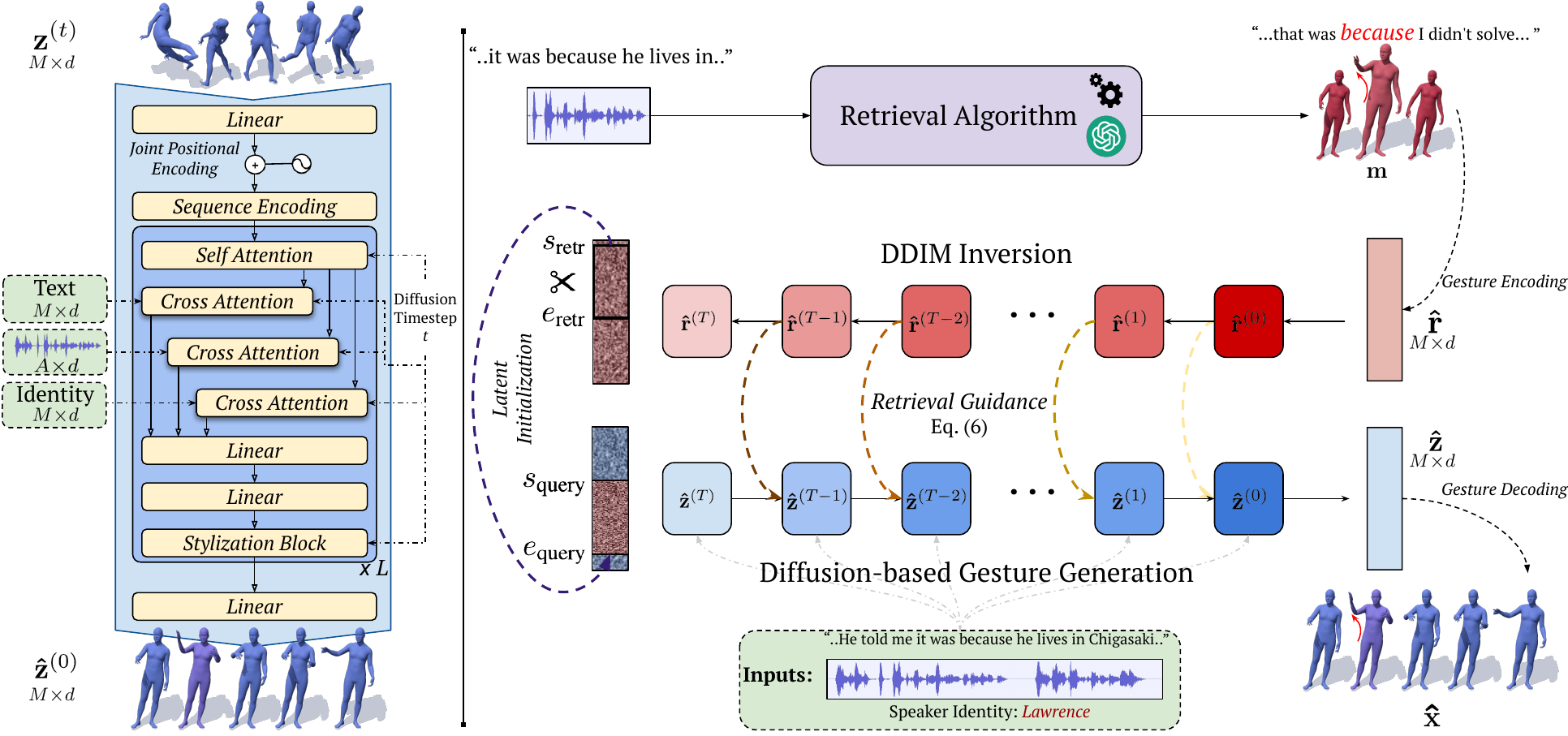}
    \caption{\textbf{\model~Framework.} 
    Our approach leverages a \textcolor{blue}{diffusion model} which predicts clean sample $\mathbf{\hat{z}}^{(0)}$ from noisy gesture sample $\mathbf{z}^{(t)}$. We then utilize retrieval algorithms~(\cref{subsec:retrieval-algos}) to modify the gesture sampling at inference time by inserting the retrieved motion through Latent Initialization (\cref{subsec:inversion}) and further controlling the sampling process through Retrieval Guidance (\cref{subsec:guidance}). This results in a \textcolor{violet}{sampled motion} which follows the \textcolor{red}{semantic retrieval}.
    \vspace{-10pt}
    }
    \label{fig:approach}
\end{figure*}
\section{Approach}
\label{sec:method}
Our primary goal is to generate semantically meaningful co-speech gesture sequences that are consistent with the content of the corresponding speech.
This goal can be partly achieved by our latent-diffusion based gesture generation model, which is trained to synthesize gestures while being conditioned on the corresponding modalities (refer to~\cref{subsec:base-model}).
This data-driven gesture generation approach is capable of generating natural looking and beat-aligned gesture sequences from the input speech.
\par
However, this basic generation framework is not sufficient to exhibit rich semantics in the generated gestures. 
Therefore, we propose additional steps that ensure the solution to two key problems --  \textit{animation} (refer to~\cref{subsec:inversion} and \cref{subsec:guidance}) and \textit{specification} (refer to~\cref{subsec:retrieval-algos}) of semantically rich gestures~\cite{neff2016hand}.
Our approach is illustrated in~\cref{fig:approach}.
\subsection{Gesture Generation with Latent-Diffusion}
\label{subsec:base-model}
As the first step, we build a robust generation framework based on latent-diffusion modeling~\cite{rombach2021ldm}, which serves as our base model to generate plausible co-speech gesture sequences.
The gesture sequence $\mathbf{x} \in \mathbb{R}^{N \times d_\mathbf{x}}$ consists of $N$ frames of human motion with $d_\mathbf{x}$ representing a combination of $J$ full body joints represented using the 6D rotation representation of~\cite{zhou2019continuity}, FLAME model parameters for face motion, root translation and foot contact labels~\cite{liu2024emage}.
Instead of training our diffusion framework on raw motion $\mathbf{x}$, we encode motion using VAE encoders, which provide separate encodings for different body parts~\cite{mughal2024convofusion}.
These part-wise VAE encodings are then used to train our conditional latent-diffusion model that generates co-speech gestures.
\paragraph{Decoupled Gesture Encoding.}
Based on the observations that each body region has different relations with speech~\cite{liu2024emage} and scale differences between them affect the generation quality~\cite{mughal2024convofusion}, we decouple the gesture sequence $\mathbf{x}$ into four different body regions: 
$\mathbf{x}_\mathbf{u} \in \mathbb{R}^{N \times 6J_u}$, 
$\mathbf{x}_\mathbf{h} \in \mathbb{R}^{N \times 6J_h}$,
$\mathbf{x}_\mathbf{f} \in \mathbb{R}^{N \times 100}$ and
$\mathbf{x}_\mathbf{l} \in \mathbb{R}^{N \times (6J_l+3+4)}$
for upper body, hands, face and a combined representation for lower body and translation respectively.
We train separate time-aware VAEs~\cite{mughal2024convofusion} for each body region, which can encode and decode the motion for each region: $\mathbf{z}_i = \xi_i(\mathbf{x}_i),~{\mathbf{x}}_i^{\prime} = \mathcal{D}_i(\mathbf{z}_i)$.
These VAEs are trained using a combination of standard reconstruction/geometric losses for motion and the KL-Divergence loss for the VAE latent space (see supplemental).
The resulting gesture representation is a set of decoupled encodings: $\mathbf{z} = \{ \mathbf{z}_\mathbf{u}, \mathbf{z}_\mathbf{h}, \mathbf{z}_\mathbf{f}, \mathbf{z}_\mathbf{l} \}$. 
This set can be concatenated to form a gesture representation $\mathbf{z} \in \mathbb{R}^{M \times d_\mathbf{z}}$, where $M<N$ is the length of time-compressed encoding corresponding to each $\mathbf{x}_i$ and $d_\mathbf{z}$ is embedding length of the representation.
Refer to the supplemental for more details.
\paragraph{Conditional Diffusion for Gesture Generation.}
Once we have obtained $\mathbf{z}$ as our gesture representation, we can frame the gesture synthesis task as that of conditional diffusion~\cite{ho2020ddpm, rombach2021ldm}, where the conditioning set~$\mathbf{C}$ comprises of audio, text and speaker embeddings. 
In this framework, the \textit{forward diffusion} process consists of Markovian chain
of successive noising steps, where Gaussian noise $\bm{\epsilon}$ is added to the encoded gesture $\mathbf{z}^{(0)}$ for $T$ timesteps until $\mathbf{z}^{(T)} \sim \mathcal{N}(0, \mathbf{I})$.
For gesture generation, the \textit{reverse process} is employed which iteratively denoises $\mathbf{z}^{(T)}$ to generate the gesture representation $\mathbf{\hat{z}}^{(0)}$. 
The process utilizes a neural network $f_{\theta}(\mathbf{z}^{(t)}, t, \mathbf{C})$ for denoising, that is trained to predict clean state~$\mathbf{\hat{z}}^{(0)}$ with the following objective:
\begin{align}
    \min_\theta \mathbb{E}_{t \sim [1,T] , {\mathbf{z}^{(0)}}, \bm{\epsilon} } || \mathbf{z}^{(0)} &- f_{\theta}(\mathbf{z}^{(t)}, t, \mathbf{C}) ||^2_2 
\end{align}
During inference, we utilize DDIM~\cite{song2021ddim} sampling for efficient generation of the gesture representation i.e. 
we generate a sample $\mathbf{\hat{z}}^{(t-1)}$ from $\mathbf{\hat{z}}^{(t)}$ using:
\begin{equation}
    \mathbf{\hat{z}}^{(t-1)} = \sqrt{\bar{\alpha}_{t-1}} \mathbf{\hat{z}}^{(0)} + \sqrt{1 - \bar{\alpha}_{t-1} - \sigma^2_{t}}~\bm{\epsilon}_\theta \big(\mathbf{\hat{z}}^{(t)}\big) + \sigma_{t}\bm{\epsilon}^{(t)}
    \label{eq:ddim-sample}
\end{equation}
Here $\bar{\alpha_t}$ controls rate of diffusion and
$\sigma_{t}$ controls stochasticity such that $\sigma_{t}=0$ results in deterministic sampling.
\par
Our denoising network consists of a transformer decoder network with $L$ decoder layers containing a separate cross-attention head for each modality from the conditioning set~$\mathbf{C}$.
The resulting activations from each head are combined using a linear layer in each decoder layer.
Diffusion timestep $t$ is passed to the decoder layers through Stylization block~\cite{zhang2024motiondiffuse} after every self-attention, cross-attention and linear layer.
We encode speech signal using wav2vec2~\cite{baevski2020wav2vec} and construct frame-aligned representation for transcription using BERT embeddings for each word~\cite{kenton2019bert}. 
Additionally, we create identity-specific speaker embeddings for input.
\subsection{Latent Initialization through DDIM Inversion}
\label{subsec:inversion}
So far, our base neural generation framework is purely data-driven and lies on one end of neural-to-explicit spectrum. 
It resembles the majority of state-of-the-art gesture generation frameworks~\cite{diffugesture, ng2024audio2photoreal, ao2023gesturediffuclip}, which rely on learned patterns to sample relevant gestures for corresponding speech and struggle to sample the semantic ones. 
On the other hand of the spectrum, pure retrieval based approaches can \naively~``paste'' raw exemplar motions during the generation process which hurts the naturalness of motion.
For instance, one could simply add noise to the retrieval and generate the rest through diffusion-based sampling~\cite{lugmayr2022diffinpaint}, which forces the generation to exactly follow the retrieval, achieving sub-par results (see~\cref{subsec:ablation} for analysis).
Therefore, we carefully infuse the retrieved gestures into the inference process by performing the retrieval transfer in the latent space of the trained diffusion model and provide a better sampling path for its generation through retrieval guidance (\cref{subsec:guidance}).
This preserves the quality of the base model, adapts it to generate semantically rich gestures and provides further control over the influence of retrieval insertion.
\par
In our framework, all retrieval algorithms (discussed later in~\cref{subsec:retrieval-algos}) must output an exemplar motion $\mathbf{m}$, which can be encoded as the \textit{retrieval} $\mathbf{r} \in \mathbb{R}^{M \times d}$ through our gesture encoder (\cref{subsec:base-model}).
We also have access to the text and audio corresponding to the retrieved motion.
Our goal is to extract the motion chunk for the retrieved semantic gesture around a word of interest (e.g. `because' in~\cref{fig:approach}) and insert it  into the motion around the corresponding words in the input text.
Specifically, let the retrieved motion contain the semantically relevant gesture chunk in the frame window $(s_{\text{retr}}, e_{\text{retr}})$ for the marked word.
The gesture in this window must be transferred onto the corresponding word location $(s_{\text{query}}, e_{\text{query}})$ in the to-be-generated sequence, also referred to as \textit{query} $\mathbf{\hat{z}}$ (see~\cref{fig:approach}).
\par
Here, the challenge is \textit{how} to insert the retrieved gesture chunk such that it is seamlessly integrated into the final inference output in terms of temporal and semantic alignment.
We perform this insertion through a combination of DDIM~\cite{song2021ddim, dhariwal2021diffusion, mokady2023null} inversion based latent initialization and a retrieval guidance objective.
DDIM inversion lets us represent the retrieval $\mathbf{r}$ into the latent space of our base diffusion model, which we then use to initialize our generation.
\par
By reversing~\cref{eq:ddim-sample} and keeping $\sigma_{t}=0$ for deterministic sampling path, we can invert $\mathbf{r}^{(0)}$ as follows:
\begin{equation}
    \mathbf{\hat{r}}^{(t+1)} = \sqrt{\bar{\alpha}_{t+1}} \mathbf{\hat{r}}^{(0)} + \sqrt{1 - \bar{\alpha}_{t+1}}\bm{\epsilon}_{\theta} \big(\mathbf{\hat{r}}^{(t)}\big)
    \label{eq:ddim-invert}
\end{equation}
This iterative process inverts the clean $\mathbf{r}^{(0)}$ to full noise $\mathbf{\hat{r}}^{(T)}$, which represents retrieved gesture in the diffusion latent space and provides the initial starting point for DDIM sampling such that it ends up at $\mathbf{\hat{r}}^{(0)} \approx \mathbf{r}^{(0)}$.
\par
To transfer the retrieved gesture onto the query gesture generation, we slice the time axis of the latent $\mathbf{\hat{r}}^{(T)}$ to extract the relevant parts using the retrieval window $[s_{\text{retr}}:e_{\text{retr}}]$ and then, place these parts onto the query window in $\mathbf{\hat{z}}^{(T)}$.
\begin{equation}
    \mathbf{\hat{z}}^{(T)} {[ s_{\text{query}} : e_{\text{query}} ]} \leftarrow \mathbf{\hat{r}}^{(T)} {[ s_{\text{retr}} : e_{\text{retr}} ]}
\end{equation}
The retrieval augmented $\mathbf{\hat{z}}^{(T)}$ can be represented as $\mathbf{\hat{z}}_\text{retr}^{(T)}$.
\begin{figure*}
	\centering
	\begin{subfigure}{0.9\linewidth}
    \centering
\includegraphics[width=\linewidth]{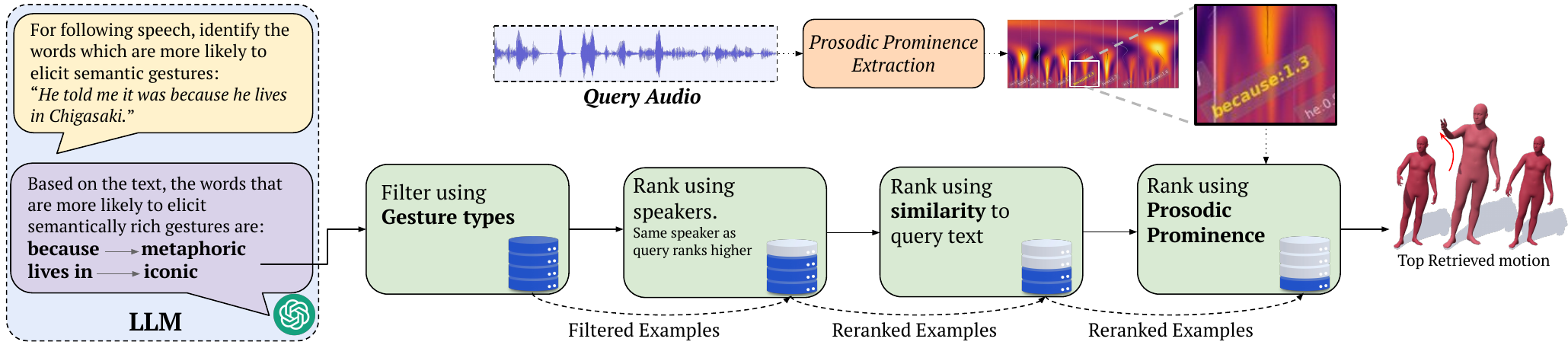}
		\caption{LLM-driven Gesture Type Retrieval}
		\label{subfig:llmgesturetype}
	\end{subfigure}
    \vfill
    \hrule
    \vspace{2pt}
	\begin{subfigure}{\linewidth}
        \centering
		\includegraphics[width=0.9\linewidth]{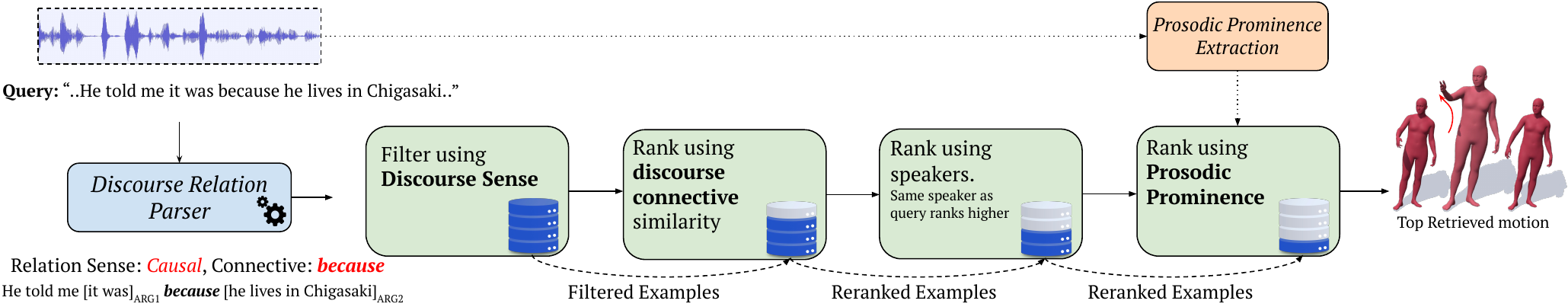}
		\caption{Discourse based Retrieval}
		\label{subfig:discourse}		
	\end{subfigure}
    % \vspace{-2pt}
	\caption{\textbf{Retrieval Algorithms.} Each algorithm parses the relevant semantic information (gesture types from LLM or discourse relations) and extracts gestures from a database by filtering examples using that information. Moreover, it also considers textual and prosodic context.
    }
	\label{fig:retrieval-algos}
    \vspace{-10pt}
\end{figure*}
\subsection{Retrieval Guidance}
\label{subsec:guidance}
As a result of latent initialization step, $\mathbf{\hat{z}}_\text{retr}^{(T)}$ comes out to be the starting point for gesture sampling process, which can be used to generate a sequence that follows the retrieved semantic gesture around the marked word.
However, there is still no control over how strongly the retrieval is followed and how the diffusion sampling process goes from $\mathbf{\hat{z}}_\text{retr}^{(T)}$ to $\mathbf{\hat{z}}_\text{retr}^{(0)}$.
Therefore, we present \textit{Retrieval Guidance} mechanism, that provides modulation control over the retrieval-augmented diffusion generation process from $\mathbf{\hat{z}}_\text{retr}^{(T)}$ to $\mathbf{\hat{z}}_\text{retr}^{(0)}$.
We utilize the sequence of inverted diffusion latents $(\mathbf{\hat{r}}^{(0)}, \mathbf{\hat{r}}^{(1)}, \dots, \mathbf{\hat{r}}^{(T-1)}, \mathbf{\hat{r}}^{(T)})$ and form our guidance objective. 
At a given diffusion timestep $t$, guidance objective and its corresponding latent update are as follows:
\begin{align}
    G_{retrieval} &= \norm{ \mathbf{\hat{z}}_\text{retr}^{(t)} {[ s_{\text{query}} : e_{\text{query}} ]} - \mathbf{\hat{r}}^{(t)} {[ s_{\text{retr}} : e_{\text{retr}} ]} }^2_2\\
    \mathbf{\Tilde{z}}_\text{retr}^{(t)} &\leftarrow \mathbf{\hat{z}}_\text{retr}^{(t)} - \lambda \nabla_{\mathbf{\hat{z}}_\text{retr}^{(t)}} G_{retrieval}
\end{align}
The updated $\mathbf{\Tilde{z}}_\text{retr}^{(t)}$ can be used further in the diffusion sampling by acting as input to the denoising network $f_{\theta}(\mathbf{\Tilde{z}}^{(t)}, t, \mathbf{C})$.
This guidance paradigm ensures that latent $\mathbf{\hat{z}}_\text{retr}^{(t)}$ follows $\mathbf{\hat{r}}^{(t)}$ at the generation timestep $t$.
\par
To control the amount of guidance at each step, one can control the number of iterations for latent update at each diffusion timestep $t$.
As an example, the generation from $\mathbf{\hat{z}}_\text{retr}^{(T)}$ without any guidance will be the weakest form of retrieval augmentation where gestures are constrained to match only at the starting timestep $t=T$.
On the other hand, if one were to make sure retrieval is followed to its strongest extent, $\mathbf{\Tilde{z}}_\text{retr}^{(t)}$ can be updated at each diffusion timestep $t$ such that $G_{retrieval} \rightarrow 0$.
Similarly, one can perform more latent updates when $t$ is closer to $T$ and lesser or no updates when $t \rightarrow 0$, in order to only perform guidance at the start of the sampling process.
Refer to~\cref{subsec:ablation} for analysis.
\subsection{Retrieval for Semantic Gesture Extraction}
\label{subsec:retrieval-algos}
Having discussed \textit{how} to insert a retrieved gesture chunk into the gesture generation process, we now discuss retrieval algorithms that utilize linguistic context to determine \textit{which} words in the query utterance are likely to invoke semantic gestures and \textit{what} are those gestures.
These algorithms retrieve semantically relevant gestures using domain knowledge of language and gesture type information.
However, our approach is not limited to these algorithms and it can be extended using any retrieval paradigm for gestures.
\par
At the core, each retrieval algorithm ranks examples in a database (the BEAT2 dataset~\cite{liu2024emage} in our case) according to a specific focus.
During ranking, some of the steps are common across algorithms and are driven by over-arching themes in gesture understanding. 
For example, motion exemplars from the same speaker typically have less speaker-dependent variation and should be ranked higher. 
Another common step across algorithms is ranking based on prosodic prominence extraction~\cite{fernandez2014gesture}. Certain elements of speech typically stand out due to variations in pitch, loudness, or duration, referred to as \textit{prosodic prominence} ~\cite{wagner2010experimental, terken2000perception}. Prosodic prominence and gestures are closely connected: co-speech gestures often align with prosodically stressed parts of speech~\cite{rohrer2023visualizing}. 
Therefore, we identify utterances in the database that match the prominence values of the marked word in the input speech~\cite{suni2017hierarchical}.
%
% 
% \par
An overview of individual steps is provided in~\cref{fig:retrieval-algos}.
% 
% \par
% 
\paragraph{LLM-based Gesture Type Retrieval.}
One of the commonly used classification for semantic gestures divides them into \textit{iconic}, \textit{metaphoric} and \textit{deictic} gestures~\cite{mcneill1992handmind}.
Kucherenko~\etal~\cite{kucherenko2021speech2properties2gestures} demonstrate the benefit of utilizing these gesture properties in the synthesis framework, as gesture type information can decouple the specification problem from the animation. 
Therefore, we leverage the reasoning abilities of an LLM~\cite{openai2024gpt} and prompt it to identify words which may elicit a semantically meaningful gesture. 
We also use it to predict the gesture type on each predicted word.
This results in a word-to-gesture type mapping, which we can use to retrieve examples from the database that contains type labels.
Details on prompts and LLM are given in the supplemental material. 
\par
Given the word-to-gesture type mapping, we take an identified word and its corresponding gesture type and filter examples in the database based on gesture type.
We leverage the semantic gesture type annotations in the BEAT2 dataset~\cite{liu2024emage} for this task. 
We then perform ranking based on speaker similarity, followed by ranking according to the text feature similarity between the query and database examples in order to match the semantic context.
Lastly, we re-rank the top semantically similar examples based on prosodic prominence values.
\cref{subfig:llmgesturetype} illustrates the steps involved.
\paragraph{Discourse-based Retrieval.}
A text has a structure that organizes its information into a coherent flow. This structure is achieved through \textit{discourse relations}, semantic-pragmatic links such as \textsc{cause-consequence} and \textsc{contrast}, which hold between clauses and sentences \citep{mann1988rhetorical,sanders1992}.
These relations can be signaled by lexical cues referred to as connectives, such as \textit{on the other hand} or \textit{because}.   
For example, in an utterance from BEAT2 dataset~\cite{liu2024emage}: \textit{``I'll go shopping \textbf{if} I'm not that tired"}, the word \textit{``if"} marks the conditional relation between two sentence segments.
Prior works have demonstrated that discourse connectives have an effect on the gestural patterns~\cite{mcneill2014discourse, calbris2011elements}.
\par
We leverage connectives to retrieve the relevant semantic gestures that co-occur with specific discourse relations.
However, connectives can be ambiguous. For example, the connective \textit{since} can signal a causal relation or a temporal one. 
We therefore retrieve the gesture co-occurring with a  connective with the same meaning, but not  necessarily the same word. For example, we can retrieve a gesture co-occurring with \textit{because} to inform the generation of a gesture co-occurring with a causal \textit{since}.
Additionally, it allows us to retrieve more samples for less frequent connectives. 
\par
In sum, we find examples in the database that carry the same discourse relational sense as the query and rank again based on the word similarity of the connective.
We conclude the algorithm with speaker-based and prosodic similarity ranking.
The steps are shown in~\cref{subfig:discourse}.
% 
% 
% 
% % 
\begin{table*}[]
\centering
\resizebox{0.9\textwidth}{!}{%
\begin{tabular}{lccccccccc}
\hline
 &
  \multicolumn{4}{c}{1 Speaker} &
   &
  \multicolumn{4}{c}{All Speakers} \\ \cline{2-5} \cline{7-10} 
 &
  FID$\downarrow$ &
  BeatAlign$\rightarrow$ &
  L1Div$\rightarrow$ &
  Diversity$\rightarrow$ &
   &
  FID$\downarrow$ &
  BeatAlign$\rightarrow$ &
  L1Div$\rightarrow$ &
  Diversity$\rightarrow$ \\ \cline{2-5} \cline{7-10} 
GT &
   &
  $0.703$ &
  $11.97$ &
  $127$ &
   &
   &
  $0.477$ &
  $7.29$ &
  $110$ \\ \hline
CaMN~\cite{liu2022beat} &
  \silver{$0.604$} &
  \gold{$0.711$} &
  $9.97$ &
  $107$ &
   &
  $0.512$ &
  $0.200$ &
  $5.58$ &
  $98$ \\
EMAGE~\cite{liu2024emage} &
  \gold{$0.570$} &
  $0.793$ &
  $11.41$ &
  $124$ &
   &
  $0.692$ &
  $0.284$ &
  $6.06$ &
  $88$ \\
Audio2Photoreal~\cite{ng2024audio2photoreal} &
  $1.02$ &
  $0.550$ &
  $12.47$ &
  $145$ &
   &
  $0.849$ &
  $0.326$ &
  $6.24$ &
  $99$ \\
ReMoDiffuse~\cite{zhang2023remodiffuse} &
  $0.702$ &
  $0.824$ &
  \silver{$12.46$} &
  {$123$} &
   &
  $1.120$ &
  $0.218$ &
  $5.06$ &
  {$116$} \\ \hline

Ours (w/ No RAG) & $
    0.911$ & 
    \silver{$0.727$} & 
    $12.78$ & 
    \silver{$130$} &
     &
    $0.519$ & 
    \silver{$0.447$} & 
    \gold{$8.64$} &
    \gold{$112$}
    \\ 
Ours (w/ Discourse) &
  $0.879$ &
  {$0.730$} &
  $12.62$ &
  \gold{$129$} &
   &
  \gold{$0.447$} &
  \gold{$0.471$} &
  \silver{$9.03$} &
  \silver{$114$} \\
Ours (w/ LLM \& Gesture Type) &
  $0.808$ &
  $0.734$ &
  \gold{$11.97$} &
  $122$ &
   &
  \silver{$0.487$} &
  {$0.514$} &
  {$9.94$} &
  $118$ \\ \hline
\end{tabular}%
}
\caption{\textbf{Comparison with state-of-the-art methods trained on BEAT2.} We demonstrate superior performance, especially when generalizing across multiple speaker identities.}
\label{tab:comparison}
\vspace{-10pt}
\end{table*}
\section{Experiments}
\label{sec:experiments}
We first compare our approach with neural state-of-the-art baselines through quantitative (\cref{subsec:metrics}) and perceptual evaluation (\cref{subsec:userstudy}).
Further, we demonstrate how our RAG approach improves upon the training-based RAG baselines.
Lastly, we validate our method design through extensive ablations and user studies (\cref{subsec:ablation}).
\par
Specifically, we compare with recent data-driven gesture generation approaches that are CaMN~\cite{liu2022beat} (LSTM-based), EMAGE~\cite{liu2024emage} (transformer-based) and Audio2Photoreal~\cite{ng2024audio2photoreal} (diffusion-based).
Furthermore, to validate RAG performance for gesture synthesis, we not only compare our method with ReMoDiffuse~\cite{zhang2023remodiffuse} (re-trained for gesture synthesis), but we also analyze it further by tailoring their approach and training it with our retrieval algorithms.
We provide seed motions for baselines~\cite{liu2024emage, liu2022beat}, but do not use seed motions to generate our results.
\paragraph{Evaluation Dataset.}
We evaluate our performance on train/val/test split from BEAT2 dataset~\cite{liu2024emage}, which contains 25 speakers.
Unlike baseline approaches for BEAT2 which report results only for a set of speakers, we perform evaluation on the test set of one speaker (\textit{Scott}) and all speakers in order to evaluate performance on large-scale multi-speaker data.
Our test set contains 265 utterances chunked into 10-sec sequences.
We retrain methods on BEAT2 if needed.
\subsection{Quantitative Evaluation}
\label{subsec:metrics}
We evaluate our method on established metrics: Beat-Alignment~\cite{aichoreo}, FID~\cite{yoon2019robots}, L1 Divergence and Diversity, which measure different aspects of the motion quality.
We tabulate the quantitative results in~\cref{tab:comparison}, where the results are divided on the basis of number of speakers used for training and evaluation.
We observe that our approach achieves state-of-the-art metric performance when trained with all speakers.
This also entails that our RAG-based diffusion approach generalizes and scales well across large-scale data with multiple speaker identities.
For the case of single speaker training, we observe best performance in terms of diversity and L1 Divergence and second-best in Beat Alignment score.
However, our diffusion-based method has higher FID compared to transformer and LSTM based methods~\cite{liu2024emage, liu2022beat} that utilize ground-truth seed and can be over-fitted to a small-scale data of single speaker, while not being able to generalize to all speakers.
\subsection{Perceptual Evaluation}
\label{subsec:userstudy}
\begin{figure}
     \centering
     \begin{subfigure}[b]{0.49\linewidth}
         \centering
         \includegraphics[width=\textwidth]{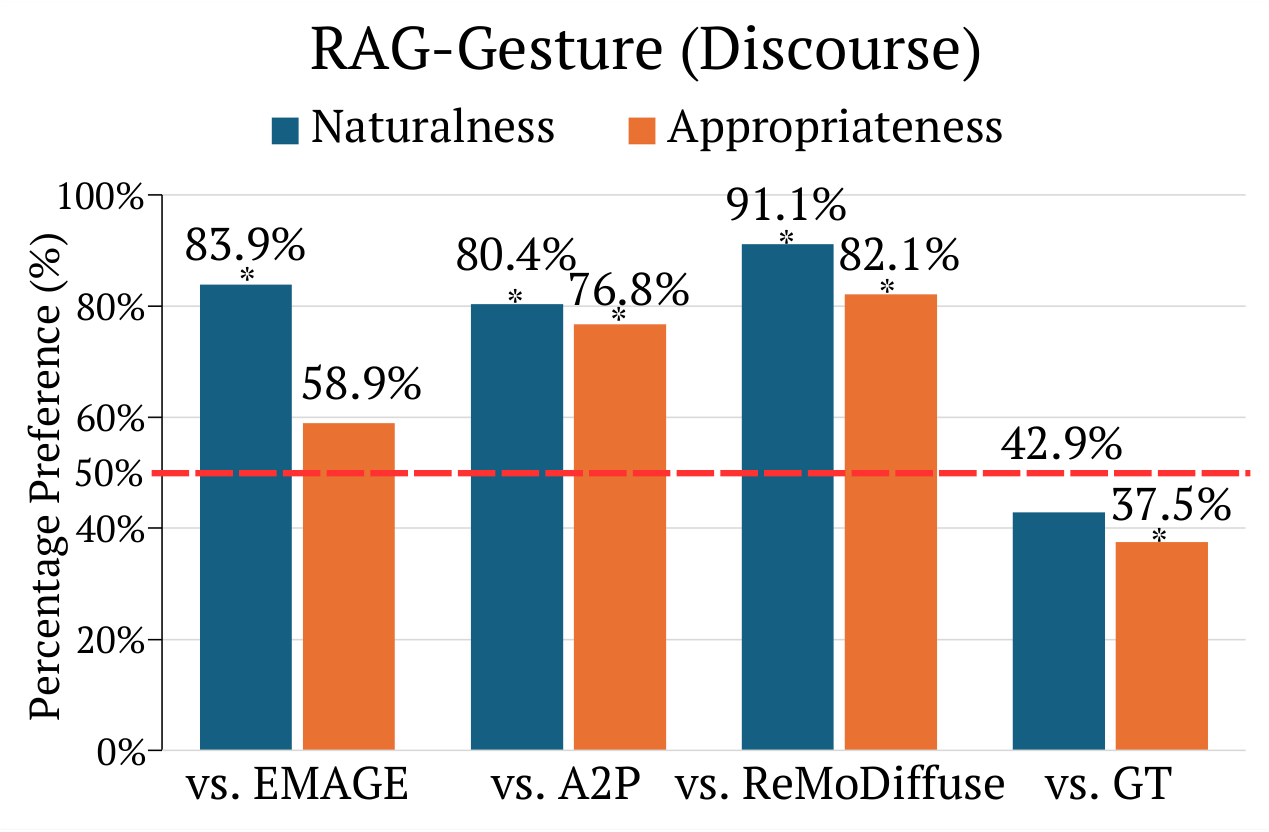}
         \label{subfig:discourse-userstudy}
         \vspace{-18pt}
     \end{subfigure}
     % \hfill
     % \qquad
     \begin{subfigure}[b]{0.49\linewidth}
         \centering
         \includegraphics[width=\textwidth]{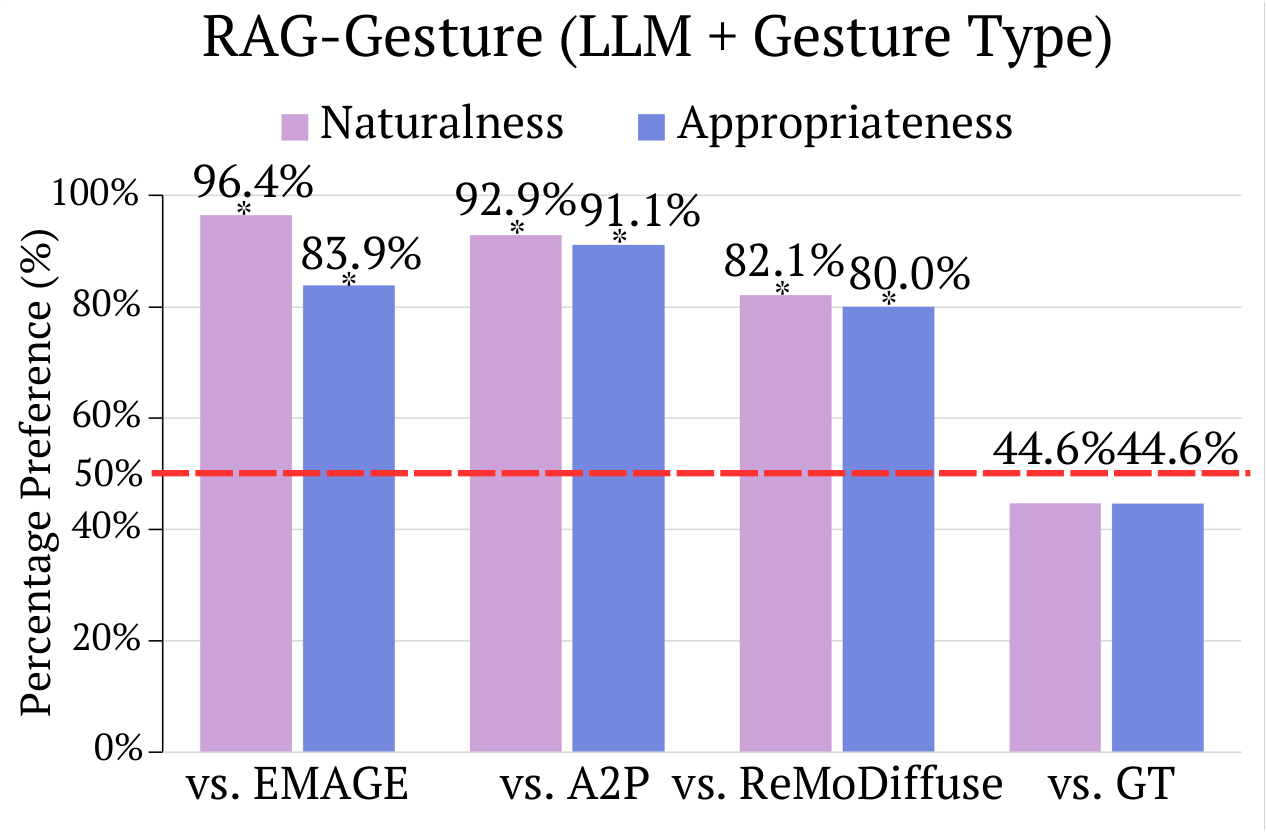}
         \label{subfig:llm-userstudy}
         \vspace{-18pt}
     \end{subfigure}
        \caption{\textbf{Results of Perceptual Evaluation.} A2P: Audio2Photoreal~\cite{ng2024audio2photoreal} GT: Ground Truth.~* denotes p-value~$<0.05$}
        \label{fig:comparison-userstudy}
        
\end{figure}
Evaluating gesture synthesis performance is a highly challenging task due to the stochasticity in gesture production and subjective nature of gesture perception~\cite{wolfert2022review, nyatsanga2023comprehensive}.
Therefore, we perform four different user studies for comparison and ablative analysis (\cref{subsec:ablation}).
First, we evaluate the synthesis quality compared to the baseline methods.
We investigate the quality of gestures based on well-established measures of \textit{naturalness} and \textit{appropriateness} in relation to speech~\cite{nagy2024towards}. 
We ask the participants to perform pairwise comparisons between generations from baselines and our method.
Results compare \model~with discourse and LLM-based retrievals against state-of-the-art methods (\cref{fig:comparison-userstudy}). 
We observe that our method achieves better perceptual quality for both naturalness and appropriateness.
For LLM based approach, we are marginally below the ground-truth preference, underscoring that our method produces not only natural, but also highly appropriate gestures. 
\subsection{Ablative Analysis}
\label{subsec:ablation}
\paragraph{Improvement over RAG Baselines.}
ReMoDiffuse~\cite{zhang2023remodiffuse} uses global text similarity for retrieval and trains the diffusion network to follow it. 
In contrast, we design our retrieval to match local context around the semantically important word and do not require training/fine-tuning for RAG unlike existing approaches~\cite{zhang2024semanticgesture,zhang2023remodiffuse}. 

Besides comparing these differences with ReMoDiffuse, we also evaluate \model~(Discourse) against a training-based approach where we incorporate our discourse-based retrievals with ReMoDiffuse's generator through the local retrieval merging strategy of SemanticGesticulator~\cite{zhang2024semanticgesture} (see supplementary for details).
Note that this architecture is trained with encoded gestures instead of raw motion and is similar to SemanticGesticulator~\cite{zhang2024semanticgesture} at a higher level, which also performs retrieval merging with encoded gestures during fine-tuning. 
\par
\begin{figure}
    \centering
     \begin{subfigure}[b]{0.49\linewidth}
         \centering
         \includegraphics[width=\textwidth]{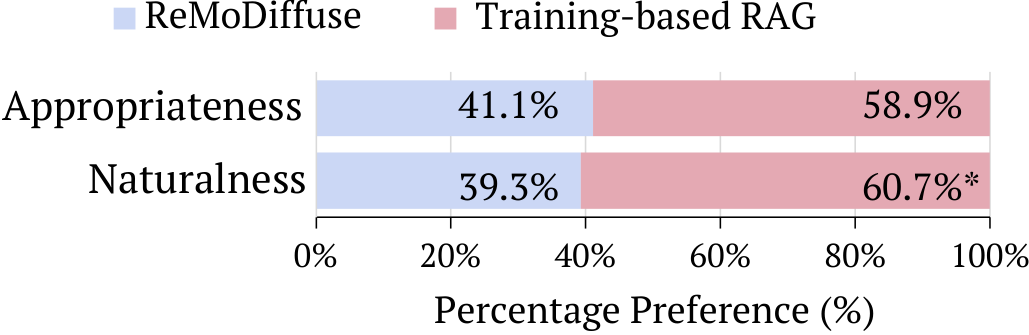}
         \label{subfig:remoret-vs-trained}
         \vspace{-18pt}
     \end{subfigure}
     % \hfill
     % \qquad
     \begin{subfigure}[b]{0.49\linewidth}
         \centering
         \includegraphics[width=\textwidth]{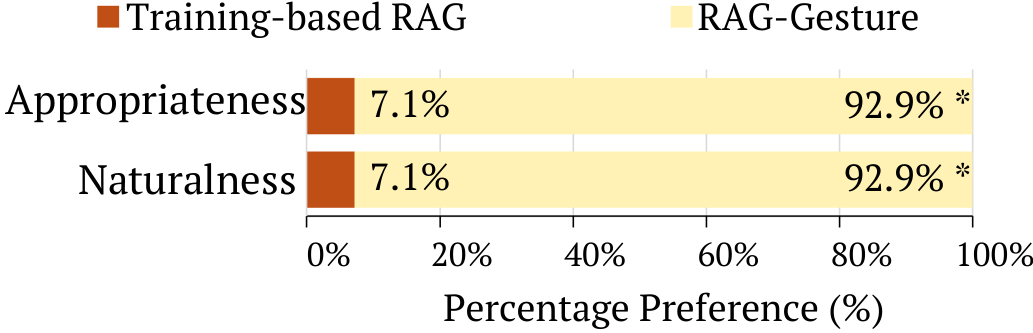}
         \label{subfig:trained-vs-inference}
         \vspace{-18pt}
     \end{subfigure}
        \caption{\textbf{Perceptual Comparison} between training-based baselines and our inference-time approach.~*: p-value~$<0.05$}
        \label{fig:remodiffuse-comp}
\end{figure}
\begin{figure}
    \centering
    \includegraphics[width=0.9\linewidth]{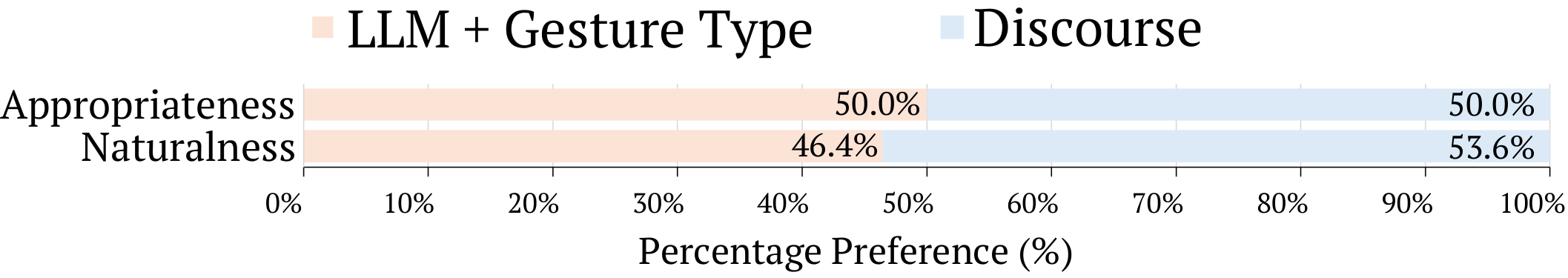}
    \caption{User Study comparing \textbf{LLM and Discourse} algorithms.}
    \label{fig:llm-v-discourse}
    \vspace{-10pt}
\end{figure}
\begin{figure}
    \centering
    \includegraphics[width=0.75\linewidth]{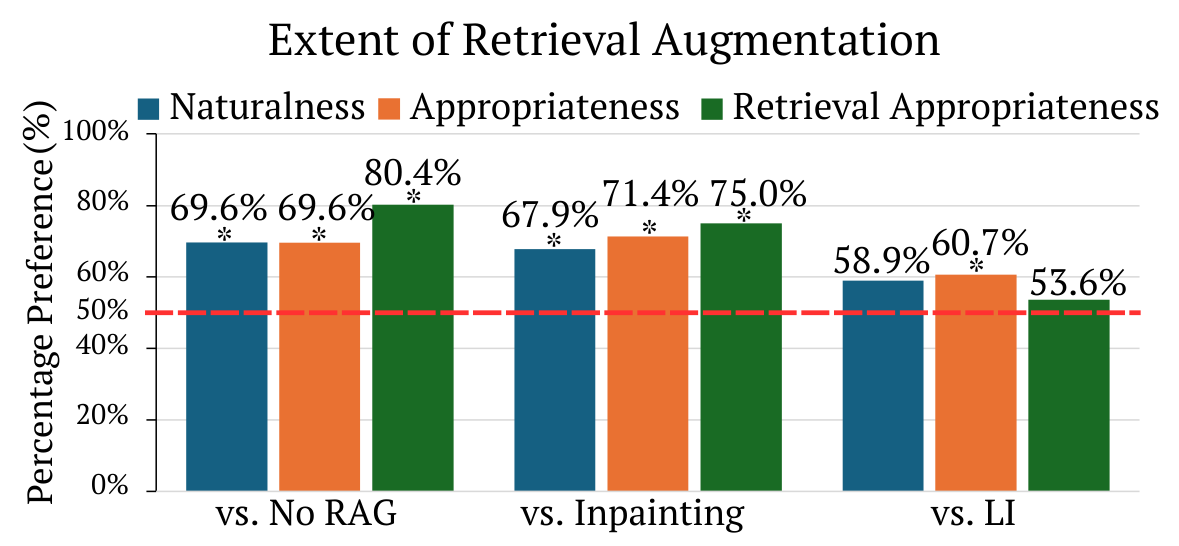}
    \caption{\textbf{Ablative Study} which illustrates preference of \model~over varying degrees of insertions.~*: p-value~$<0.05$}
    \label{fig:extent-retrievalinjection}
    \vspace{-10pt}
\end{figure}

We evaluate this quantitatively in~\cref{tab:remodiffuse-comp} and perform an additional user study (\cref{fig:remodiffuse-comp}). 
Moreover, we also report MPJPE between the generated gesture and retrieved motion during the retrieval insertion window to measure the retrieval following performance.
\begin{table}[]
\centering
\resizebox{\columnwidth}{!}{%
\begin{tabular}{lcccc}
\hline
                           & FID$\downarrow$     & BeatAlign$\rightarrow$ & L1Div$\rightarrow$   & MPJPE (mm)$\downarrow$ \\ \hline
GT                         &         & $0.477$   & $7.29$ &            \\ \hline
ReMoDiffuse~\cite{zhang2023remodiffuse}                 & $1.120$ & $0.218$ & $5.06$  & $200.5$ \\
Training-based RAG & $0.525$ & $0.414$ & $6.91$ & $34.6$  \\
\model~ & $0.447$ & $0.471$   & $9.03$ & $35.0$ \\   \hline
\end{tabular}%
}
\caption{Quantitative Comparison between RAG baselines.}
\label{tab:remodiffuse-comp}
\vspace{-10pt}
\end{table}
Firstly, we observe that the training-based RAG which uses local semantic retrieval during training performs better than ReMoDiffuse, both in terms of perceptual quality and metrics.
Moreover, we also see higher faithfulness to the retrievals through low MPJPE which entails that properly specifying the retrieved motion on the exact word affects the gesture quality.
Secondly, our inference-time \model~is better than the training-based model, even after utilizing our retrieval algorithm.
This underscores our method's superiority and the importance of having a good gesture animation performance.
\paragraph{LLM-based Gesture Type vs. Discourse Retrieval.}
For evaluating perceptual differences between LLM-based gesture type retrieval and discourse-based retrieval, we conduct an additional user study where we ask the participants to evaluate differences between generated gestures from the two algorithms in terms of naturalness and appropriateness (\cref{fig:llm-v-discourse}).
Interestingly, we find that evaluators do not show any strong preference between the two algorithms.
\paragraph{Extent of Retrieval Augmentation}
Since our approach can control the influence of the retrieved gesture onto the generated one, we analyze how much influence is optimal to synthesize semantically accurate yet natural looking gestures. 
We compare \model's prediction against varying degrees of retrieval insertion (\cref{tab:insertion-extent}) using Discourse-based retrieval.
``Inpainting'' means adding noise to the retrieval and synthesizing the remaining motion and ``No RAG'' represents no retrieval insertion. 
We conduct another user study with a third ``Retrieval Appropriateness'' question to gauge semantic appropriateness of the generated gesture during the insertion window~(\cref{fig:extent-retrievalinjection}).
Interestingly, we observe increasing preference of our retrieval-guided model as we move towards both extremes of the insertion.
In terms of gesture perception, evaluators neither prefer generation without RAG (purely data-driven) nor a strong insertion through Inpainting (purely retrieval-based).
Lastly, our model achieves good balance of retrieval following in terms of MPJPE, prosodic alignment and diversity. 
\begin{table}[]
\centering
\resizebox{\columnwidth}{!}{%
\begin{tabular}{lcccc}
\hline
                                        & FID$\downarrow$     & BeatAlign$\rightarrow$ & L1Div$\rightarrow$   & MPJPE (mm)$\downarrow$ \\ \hline
GT                                      &         & $0.477$   & $7.29$  &            \\ \hline
No RAG                                  & $0.519$ & $0.447$   & $8.64$ & $59.6$     \\
RAG by Inpainting~\cite{lugmayr2022diffinpaint}                       & {$0.446$} & $0.450$   & {$8.46$} & {$28.1$}     \\
RAG by LI only                          & $0.589$ & {$0.486$}   & $9.29$ & $54.6$     \\
\model~(LI+RG)                            & {$0.447$} & {$0.471$}   & $9.04$ & {$35.0$}     \\ \hline
\end{tabular}%
}
\caption{\textbf{Extent of Semantic Retrieval Insertion.} LI: Latent Initialization; RG: Retrieval Guidance. }
\label{tab:insertion-extent}
\vspace{-10pt}
\end{table}
\section{Conclusion}
\label{sec:conclusion}
In this work, we present \model~-- a novel approach that leverages interpretable retrieval algorithms to introduce inference-time RAG to a diffusion-based gesture generator. 
Therefore, it combines advantages of the pure retrieval-based and fully data-driven approaches to synthesize natural yet semantically meaningful gestures.
Perceptual evaluation demonstrates that this type of neuro-explicit approach is better than either of the two approaches and existing state-of-the-art methods. 
Moreover, we demonstrate that LLM-driven gesture type prediction and discourse relations can enhance semantic understanding in current gesture generation architectures.
Lastly, our framework can be extended to synthesize gestures that exhibit task-specific gestural patterns like referential or emotion-specific gestures.
\paragraph{Acknowledgements.}
% \label{sec:acknowledgements}
% 
This work was funded by the Deutsche Forschungsgemeinschaft (DFG, German Research Foundation) -- GRK 2853/1 “Neuroexplicit Models of Language, Vision, and Action” - project number 471607914.
The third author of this study
(MS) was supported through the NWO-funded project ``Searching for meaning: Identifying and interpreting alternative discourse relation signals in multi-modal language comprehension'' (VI.Veni.231C.021).
{
    \small
    \bibliographystyle{ieeenat_fullname}
    \bibliography{main}
}

\clearpage
\setcounter{page}{1}
\maketitlesupplementary
We first discuss limitations of our method and provide details on the perceptual evaluation. 
Further, we elaborate on evaluation metrics and provide additional experiments and analyses. 
Lastly, we discuss our implementation and provide analysis and runtime details for it. 
\section{Limitations}
\label{sec:limitations}
Our method relies on the sparse semantic data which is extracted from the BEAT2 dataset~\cite{liu2024emage}.
This creates a scarcity of good exemplars which can be used during the database matching steps. 
As a result, the LLM-based Gesture Type algorithm sometimes struggles to find good contextual matches for each gesture type and word identified by the LLM. 
\par
Secondly, our method combines explicit rule-based algorithms with a neural generation framework. 
In rare cases, these algorithms can fail in edge cases and result in an incorrect retrieved example. 
However, our learned framework on top of the retrieval algorithms mitigates this and ignores out-of-distribution motion exemplars since it has been trained to produce only those gestures which match the speech. 
This also is affected by the extent of retrieval augmentation.
\section{Details on User Study}
For the perceptual evaluation of gesture generation capability, 56 participants were shown a randomly sampled set of 16 forced-choice questions. 
Our study consists of four sections corresponding to four different user studies.
First section focuses on the comparison with the state-of-the-art approaches.
Each question comprises of a side-by-side animation of our method along with one of EMAGE~\cite{liu2024emage}, Audio2Photoreal~\cite{ng2024audio2photoreal}, RemoDiffuse~\cite{zhang2023remodiffuse} or the ground-truth.
Second section is a short section where we compare LLM-driven gesture generation with discourse-based synthesis. 
Third, we perform pair-wise comparisons between results from one of the RAG baselines (derived from ReMoDiffuse~\cite{zhang2023remodiffuse}) and our method.
Lastly, we evaluate different approaches to perform RAG by controlling its extent.
\par
In first three sections, we try to evaluate naturalness and appropriateness.
Specifically, we ask two questions (a) \textit{``Which of the two gestures look natural?''} and (b) \textit{``Which of them looks appropriately aligned to what the person is saying?''}.
In the last section, we add an additional question which focuses on gauging the semantic appropriateness in the retrieval window, with a goal of evaluating the RAG capability.
We highlight the identified words from the retrieval algorithms and add an additional question: \textit{``Which of the two have better gestures in the highlighted section, especially at the capitalized word in the prompt?''}
\section{Evaluation Metrics}
\paragraph{FID.}
We employ the Frechet Inception Distance (FID) metric inspired by Yoon~\etal~\cite{yoon2020trimodal}, which is also known as FGD.
We use the autoencoder network provided by BEAT2~\cite{liu2024emage} to get the gesture encodings for FID evaluation and do not retrain our own network.
\paragraph{Beat Alignment Score.}
Originally introduced to measure alignment of music beats to dance motion, Beat Alignment Score~\cite{aichoreo} has been adapted for the gesture synthesis task where it aims to measure the correlation between gesture beats and audio beats.
\paragraph{L1 Divergence.}
This metric (also called L1 variance) measures the distance of all frames in a single generated sample from their mean. It is helpful in identifying synthesized gestures that are static and unexpressive.
\paragraph{Diversity.} 
It computes the average pairwise Euclidean distance between the generated gestures from the test set.
\paragraph{Multi-modality.}
This metric requires sampling different gesture motions for a single speech input from the generative model~\cite{dabral2023mofusion}. Then, it computes Euclidean distance between the diverse generated gestures. It probes the diverse sampling capabilities of a generative model. 
\begin{table*}[]
\centering
\resizebox{\textwidth}{!}{%
\begin{tabular}{lccccccccccccc}
 &
  \multicolumn{6}{c}{FID$\downarrow$} &
   &
  \multicolumn{6}{c}{Multimodality$\uparrow$} \\ \cline{2-7} \cline{9-14} 
 &
  CaMN &
  EMAGE &
  Audio2Photoreal &
  \multicolumn{1}{c|}{ReMoDiffuse} &
  Ours (w/ Discourse) &
  Ours (w/ LLM \& Gesture Type) &
   &
  CaMN &
  EMAGE &
  Audio2Photoreal &
  \multicolumn{1}{l|}{ReMoDiffuse} &
  Ours (w/ Discourse) &
  Ours (w/ LLM \& Gesture Type) \\ \hline
\multicolumn{1}{l|}{wayne} &
  \gold{1.23} &
  2.06 &
  2.32 &
  \multicolumn{1}{c|}{3.58} &
  \silver{1.49} &
  1.59 &
   &
  n/a &
  n/a &
  1.1 &
  \multicolumn{1}{l|}{\silver{3.4}} &
  3.1 &
  \gold{3.7} \\
\multicolumn{1}{l|}{scott} &
  0.83 &
  1.17 &
  1.02 &
  \multicolumn{1}{c|}{1.76} &
  \gold{0.78} &
  {0.83} &
   &
  n/a &
  n/a &
  1.5 &
  \multicolumn{1}{l|}{\gold{7.7}} &
  5.3 &
  \silver{5.4} \\
\multicolumn{1}{l|}{solomon} &
  1.22 &
  1.42 &
  1.93 &
  \multicolumn{1}{c|}{2.45} &
  \silver{0.92} &
  \gold{0.86} &
   &
  n/a &
  n/a &
  0.5 &
  \multicolumn{1}{l|}{3.5} &
  \silver{4.3} &
  \gold{4.5} \\
\multicolumn{1}{l|}{lawrence} &
  0.98 &
  1.39 &
  1.13 &
  \multicolumn{1}{c|}{3.16} &
  \silver{0.69} &
  \gold{0.66} &
   &
  n/a &
  n/a &
  1.9 &
  \multicolumn{1}{l|}{\gold{6.9}} &
  5.6 &
  \silver{6.3} \\
\multicolumn{1}{l|}{stewart} &
  \gold{0.65} &
  \silver{1.26} &
  1.62 &
  \multicolumn{1}{c|}{1.76} &
  1.49 &
  1.49 &
   &
  n/a &
  n/a &
  0.3 &
  \multicolumn{1}{l|}{0.7} &
  \silver{2.4} &
  \gold{3.0} \\
\multicolumn{1}{l|}{carla} &
  \gold{0.81} &
  1.40 &
  \silver{1.33} &
  \multicolumn{1}{c|}{2.95} &
  1.63 &
  1.49 &
   &
  n/a &
  n/a &
  0.4 &
  \multicolumn{1}{l|}{1.3} &
  \silver{1.4} &
  \gold{1.5} \\
\multicolumn{1}{l|}{sophie} &
  \gold{0.92} &
  \silver{1.67} &
  1.85 &
  \multicolumn{1}{c|}{2.76} &
  1.76 &
  1.74 &
   &
  n/a &
  n/a &
  0.69 &
  \multicolumn{1}{l|}{\gold{3.5}} &
  2.7 &
  \silver{3.2} \\
\multicolumn{1}{l|}{miranda} &
  \gold{0.58} &
  \silver{0.87} &
  1.10 &
  \multicolumn{1}{c|}{1.86} &
  1.09 &
  1.34 &
   &
  n/a &
  n/a &
  0.4 &
  \multicolumn{1}{l|}{0.9} &
  \silver{1.7} &
  \gold{1.9} \\
\multicolumn{1}{l|}{kieks} &
  \gold{1.30} &
  2.62 &
  1.90 &
  \multicolumn{1}{c|}{7.65} &
  1.74 &
  \silver{1.63} &
   &
  n/a &
  n/a &
  1.0 &
  \multicolumn{1}{l|}{2.1} &
  \silver{3.6} &
  \gold{4.1} \\
\multicolumn{1}{l|}{nidal} &
  \gold{0.40} &
  0.65 &
  0.72 &
  \multicolumn{1}{c|}{1.74} &
  0.67 &
  \silver{0.64} &
   &
  n/a &
  n/a &
  0.7 &
  \multicolumn{1}{l|}{1.9} &
  \silver{2.6} &
  \gold{3.1} \\
\multicolumn{1}{l|}{zhao} &
  1.66 &
  2.70 &
  1.96 &
  \multicolumn{1}{c|}{3.37} &
  \silver{1.37} &
  \gold{1.32} &
   &
  n/a &
  n/a &
  1.3 &
  \multicolumn{1}{l|}{3.2} &
  \silver{3.3} &
  \gold{3.4} \\
\multicolumn{1}{l|}{lu} &
  1.40 &
  2.73 &
  1.92 &
  \multicolumn{1}{c|}{2.23} &
  \silver{1.27} &
  \gold{1.16} &
   &
  n/a &
  n/a &
  0.7 &
  \multicolumn{1}{l|}{1.7} &
  \silver{2.4} &
  \gold{2.6} \\
\multicolumn{1}{l|}{carlos} &
  \gold{0.78} &
  \silver{1.47} &
  1.71 &
  \multicolumn{1}{c|}{2.47} &
  1.95 &
  2.56 &
   &
  n/a &
  n/a &
  0.2 &
  \multicolumn{1}{l|}{2.5} &
  \silver{2.9} &
  \gold{3.1} \\
\multicolumn{1}{l|}{jorge} &
  \gold{1.49} &
  2.57 &
  1.97 &
  \multicolumn{1}{c|}{3.55} &
  \silver{1.89} &
  1.93 &
   &
  n/a &
  n/a &
  0.3 &
  \multicolumn{1}{l|}{1.7} &
  \silver{1.9} &
  \gold{2.1} \\
\multicolumn{1}{l|}{itoi} &
  \gold{0.93} &
  1.61 &
  1.34 &
  \multicolumn{1}{c|}{2.28} &
  \silver{1.07} &
  1.32 &
   &
  n/a &
  n/a &
  0.8 &
  \multicolumn{1}{l|}{1.8} &
  \silver{3.0} &
  \gold{3.1} \\
\multicolumn{1}{l|}{daiki} &
  \gold{0.78} &
  1.78 &
  1.66 &
  \multicolumn{1}{c|}{3.04} &
  \silver{0.91} &
  1.19 &
   &
  n/a &
  n/a &
  0.3 &
  \multicolumn{1}{l|}{2.3} &
  2.3 &
  \gold{2.7} \\
\multicolumn{1}{l|}{li} &
  1.10 &
  1.74 &
  1.17 &
  \multicolumn{1}{c|}{2.06} &
  \gold{0.71} &
  \silver{0.79} &
   &
  n/a &
  n/a &
  0.6 &
  \multicolumn{1}{l|}{\gold{3.9}} &
  2.7 &
  \silver{3.0} \\
\multicolumn{1}{l|}{ayana} &
  \gold{1.19} &
  2.03 &
  \silver{1.96} &
  \multicolumn{1}{c|}{4.35} &
  2.09 &
  2.13 &
   &
  n/a &
  n/a &
  0.4 &
  \multicolumn{1}{l|}{{1.8}} &
  \silver{1.9} &
  \gold{2.1} \\
\multicolumn{1}{l|}{luqi} &
  \gold{1.25} &
  2.22 &
  1.67 &
  \multicolumn{1}{c|}{5.21} &
  \silver{1.38} &
  1.86 &
   &
  n/a &
  n/a &
  0.7 &
  \multicolumn{1}{l|}{{1.7}} &
  \silver{2.2} &
  \gold{2.5} \\
\multicolumn{1}{l|}{hailing} &
  \gold{0.53} &
  \silver{1.20} &
  7.53 &
  \multicolumn{1}{c|}{5.72} &
  2.35 &
  2.79 &
   &
  n/a &
  n/a &
  0.3 &
  \multicolumn{1}{l|}{1.0} &
  \silver{2.5} &
  \gold{2.6} \\
\multicolumn{1}{l|}{kexin} &
  1.07 &
  1.70 &
  1.19 &
  \multicolumn{1}{c|}{1.87} &
  \gold{0.92} &
  \silver{0.95} &
   &
  n/a &
  n/a &
  0.3 &
  \multicolumn{1}{l|}{\silver{1.8}} &
  1.6 &
  \gold{1.9} \\
\multicolumn{1}{l|}{goto} &
  \gold{0.84} &
  \silver{1.32} &
  2.01 &
  \multicolumn{1}{c|}{2.51} &
  1.45 &
  2.09 &
   &
  n/a &
  n/a &
  0.3 &
  \multicolumn{1}{l|}{\silver{2.3}} &
  2.0 &
  \gold{2.4} \\
\multicolumn{1}{l|}{yingqing} &
  \gold{1.67} &
  2.50 &
  2.00 &
  \multicolumn{1}{c|}{4.34} &
  1.82 &
  \silver{1.74} &
   &
  n/a &
  n/a &
  1.1 &
  \multicolumn{1}{l|}{\gold{3.5}} &
  3.0 &
  \silver{3.2} \\
\multicolumn{1}{l|}{tiffnay} &
  \gold{0.81} &
  1.35 &
  1.09 &
  \multicolumn{1}{c|}{2.67} &
  \silver{0.92} &
  1.11 &
   &
  n/a &
  n/a &
  0.3 &
  \multicolumn{1}{l|}{\silver{1.3}} &
  1.2 &
  \gold{1.6} \\
\multicolumn{1}{l|}{katya} &
  \gold{1.10} &
  2.09 &
  1.57 &
  \multicolumn{1}{c|}{2.65} &
  \silver{1.15} &
  1.23 &
   &
  n/a &
  n/a &
  0.6 &
  \multicolumn{1}{l|}{\gold{2.9}} &
  2.3 &
  \silver{2.6} \\ \hline
\end{tabular}%
}
\caption{Per Speaker FID/Multimodality}
\label{tab:perspk-fid-mm}
\end{table*}
\begin{table}[t]
\centering
\resizebox{\columnwidth}{!}{%
\begin{tabular}{cccccc}
\hline
\multicolumn{6}{c}{Multi-modality $\uparrow$}                                      \\ \hline
CaMN & EMAGE & Audio2Photoreal & RemoDiffuse & Ours (w/ Discourse) & Ours (w/ LLM) \\
n/a  & n/a   & 16.9            & 66.5        & \silver{69.1}       & \gold{76.7}   \\ \hline
\end{tabular}%
}
\caption{Overall Multi-modality for All-Speaker Model}
\label{tab:base-mm}
\end{table}
% % 
% 
% 
% 
\section{Additional Experiments}
\subsection{Per-speaker Quantitative Comparison with FID \& Multi-modality}
\label{subsec:perspk}
To perform a robust evaluation on speaker generalizability of our framework, we provide per-speaker FID and Multi-modality metrics for the all-speaker model in~\cref{tab:perspk-fid-mm}.
We observe that our framework achieves best FID for large number of the speakers, which shows that our method generalizes well to the speaker specific patterns and idiosyncrasies despite taking no seed gestures at input.
Overall, CaMN~\cite{liu2022beat}, achieves lower FID because it uses seed input from the ground truth data which results in lower scores. 
Due to the same reason, EMAGE~\cite{liu2024emage} also gets lower FID.
\par
However, CaMN and EMAGE always generate same gestures for a given speech input, so they perform worse in terms of Multi-modality, which makes them less ideal for diverse gesture generation. 
In contrast, our approach gets the best results showing diverse gesture generation capabilities of our model (~\cref{tab:base-mm}).
\subsection{Comparison of RAG with Motion Blending}
% 
% \begin{table}[h]
% % \vspace{-1em}
% \centering
% \resizebox{\linewidth}{!}{%
% \begin{tabular}{lccccc}
% \hline
%  & FID$\downarrow$ & BeatAlign$\rightarrow$ & L1Div$\rightarrow$ & Diversity (Div.)$\rightarrow$ & MPJPE$\downarrow$ \\ \cline{2-6} 
% GT                 &         & $0.477$ & $7.29$ & $110$ &        \\ \hline
% LinearBlending     & $0.346$ & $0.394$ & $8.74$ & $111$ & $4.6$  \\
% Ours(w/ Discourse) & $0.447$ & $0.471$ & $9.03$ & $114$ & $35.0$ \\ \hline
% \end{tabular}
% }
% % \vspace{-1.2em}
% \end{table}
Linear motion blending can be considered as an alternative for exemplar insertion in the generated motions. 
Therefore, we explore this alternative by pasting semantic examples onto the retrieval windows and blending motion at the window boundaries.
We observe smoothing artifacts with motion blending where motion looks unnatural and gesture beats are smoothened around window boundaries due to interpolation.
Compared to this, motion naturalness and speech-to-gesture alignment are preserved by using the proposed retrieval insertion method, which ``blends'' motion in the diffusion latent space.
\textit{Video results} can be seen in the supplementary video (@12:54).
\subsection{Ablation on single speaker training}
We observe high FID for the single-speaker setting. Therefore, we analyze this further in \cref{tab:1spk-abl} by comparing single-speaker model with Non-RAG version (to check underfitting), and also with an RAG version which uses larger all-speaker database but same model trained on a single speaker (to check if smaller database causes worse performance).
\begin{table}
% \vspace{-1em}
\centering
\resizebox{\columnwidth}{!}{%
\begin{tabular}{lcccc}
\hline
                        & FID$\downarrow$ & BeatAlign$\rightarrow$ & L1Div$\rightarrow$ & Diversity$\rightarrow$ \\ \cline{2-5} 
GT                      &                 & $0.703$                & $11.97$            & $127$                  \\ \hline
No RAG                  & $0.911$         & $0.727$                & $12.78$            & $130$                  \\
RAG with all-speaker DB (w/ Discourse) & $0.872$         & $0.727$                & $12.53$            & $127$                  \\
RAG with 1-speaker DB (w/ Discourse)   & $0.879$         & $0.730$                & $12.62$            & $129$                  \\ \hline
\end{tabular}
}
\caption{Quantitative Ablation with 1-speaker training.}
\label{tab:1spk-abl}
% \vspace{-1em}
\end{table}
We observe the same trend as the ablative analysis, that RAG versions of the model perform better. 
Interestingly, usage of larger DB performs slightly better due to better example matching.
However, FID remains higher than CaMN/EMAGE and metrics show little difference between 1-speaker model and ``No RAG'' model, which shows underfitting due to small data size.
Our experiments concur that diffusion models are more data-hungry and train better with larger data. Moreover, we observe that deterministic models (CaMN/EMAGE) that use seed motion, perform better with the smaller data by overfitting and predicting samples closer to GT (\cref{subsec:perspk}). 
% 
% \HM{mention tradeoff - depending on parameters, you either loose beats or motion quality. we do this latent space thing (also consistent with literature.) }
% \par
% % 
% 

\section{Implementation Details \& Analysis}
\label{suppsec:implementation}
\subsection{Input Representations}
Representing speech and its transcription is highly important aspect of diffusion-based gesture modelling process.
In our experiments, we found that changing the structure of text embeddings affects gesture understanding during the learning process, which consequently is reflected during the synthesis phase as well. 
To construct our text representation, we build a per-frame embedding with corresponding word embeddings residing on each frame.
To extract discourse connectives, we pass text transcriptions of speech samples through discopy~\cite{knaebel2021discopy} and store resulting outputs along with dataset samples.
We compute the word embeddings by aggregating sub-word token activations from last 4 layers of BERT model~\cite{kenton2019bert}.
\subsection{Decoupled gesture encoding}
We utilize time-aware VAE architecture by Mughal~\etal~\cite{mughal2024convofusion}.
This architecture utilizes seperate encoders for frame window chunks of original motion to encode each chunk into an encoding. 
Then, it jointly decodes all of the chunks together to reconstruct the original motion.
We use $N=150$ representing $10$ seconds of motion at $15$ frames per second.
Moreover, the frame chunk length of our time-aware VAE is $15$, making each chunk encoding correspond to $1$ second of motion.
This results in a chunked gesture encoding of length $10$ for each body part.
Finally, we concatenate all 4 body part encodings along the time axis with separators in between them, resulting in $M = 40 + 3 = 43$.
\par
We train the VAE on the reconstruction task by utilizing a set of losses to optimize the model.
We apply Geodesic Loss on rotation matrices and standard MSE losses on 6D, axis-angle and joint position representation of the motion. 
Moreover, we also apply additional MSE losses to optimize velocity/acceleration of motion~\cite{ghosh2022imos}.
Lastly, we apply loss on foot contact predictions during VAE training to reduce foot sliding~\cite{ghosh2024remos, zhang2024roam}.
\subsection{RAG-driven Gesture Diffusion model}
To optimize our diffusion model, we utilize Adam~\cite{diederik2014adam} with a learning rate of $1e-4$.
We utilize ``scaled linear'' as our $\beta_t$ schedule and use $1000$ steps while training. 
For inference, we use spaced $50$ steps with DDIM scheduler.
The transformer network contains 16 attention heads and 8 decoder layers.
To better disambiguate body parts in our gesture encoding, we also add a separate sinusoidal positional encoding for body parts.
\paragraph{Retrieved Motion Insertion.}
In order to insert the retrieved gestures into the query latents, we only consider latents for upper body and hands. 
The encodings of these body parts are transferred from retrieval to query sample because speech has the most amount of semantic significance on these two body regions in terms of co-verbal gestures.
\par
In the current setup, the insertion of retrieved gestures happens at $t=T$ where the latents are fully noised. However, our implementation also allows to arbitrarily choose a timestamp $t=K$ for retrieval insertion. Consequently, one can then perform Retrieval Guidance for steps $t<K$.
% \HM{add p\% inversion point }
% 
\subsection{Details on LLM Prompting}
We utilize OpenAI's gpt-4o-mini model for semantic gesture type prediction. 
We provide a system prompt containing a brief explanation of gesture types and a user prompt which contains text from the test dataset and the question.
\par
\paragraph{System Prompt.}
``You are an expert in human gestures. You need to identify words that may elicit semantically meaningful gestures(deictic, iconic, metaphoric) and their types:
(a) Metaphoric Gesture: Represents abstract ideas or concepts physically, creating a vivid mental image.
(b) Iconic Gesture: Mimics the shape or action of the object or concept being described.
(c) Deictic Gesture: Points to or indicates a person, object, or location.
Format your response as a python list of python tuples of (word, type). For example: [('hello', 'beat'), ('world',
'iconic')]''.
\paragraph{User Prompt.}
Identify at most 2 important words which are more likely to elicit semantically meaningful gestures and what are types of those gestures in following text: ``{SAMPLE TEXT}''.
\paragraph{Effect of Word Number in Prompt.} 
% 
% As \cref{suppsec:implementation} shows that 
% 
\begin{table}
    % \vspace{-1.4em}
    \centering
    % \resizebox{\linewidth}{!}{
    \begin{tabular}{lcccc}
\hline
        & FID$\downarrow$ & BeatAlign$\rightarrow$ & L1Div$\rightarrow$ & Diversity$\rightarrow$ \\ \cline{2-5} 
GT      &                 & $0.477$                & $7.29$             & $110$                  \\ \hline
1-word & $0.483$         & $0.486$                & $9.38$             & $115$                  \\
2-word  & $0.487$         & $0.514$                & $9.94$             & $118$                  \\
3-word  & $0.510$         & $0.536$                & $10.26$            & $120$                  \\ \hline
\end{tabular}
\caption{Quantitative Comparison using different quantities of identified words in LLM prompt.}
% }
% \vspace{-1.5em}
% 
\label{tab:prompt-ablation}
\end{table}
As shown above, the user prompt contains a maximum number of words for which LLM needs to predict gesture types. As this number of words is a hyperparameter, we perform quantitative comparison for different quantities in~\cref{tab:prompt-ablation}. Results show slight variation in metrics and even smaller difference in terms of perceptual quality. Therefore, we conclude that retrieval algorithm is flexible enough to be used with any configuration.
\subsection{Baseline Retraining Details}
To be consistent with single speaker evaluation on BEAT2 dataset, we utilize released model weights by EMAGE~\cite{liu2024emage}. For other approaches (including ours), we retrain the method on single speaker data belonging to the speaker \textit{``Scott''}.
Since there are no available models for the chosen baselines which have been trained on all speakers in BEAT2 dataset, we train all the methods on complete dataset through their provided codebases.
Methods which do not contain speaker specific generalizations like Audio2Photoreal~\cite{ng2024audio2photoreal}, are modified to include a speaker embedding along with text and speech embeddings.
Moreover, Audio2Photoreal is adapted to support the skeletal format of BEAT2.
% 
% \par
% 
Lastly, ReMoDiffuse, originally released for text-to-motion task, is modified for gesture synthesis and their retrieval process is implemented using text feature similarity method.
\par
For the comparison of our approach with training-based RAG (Sec. 4.3), we further modify the ReMoDiffuse architecture and train it using our gesture encodings instead of raw motion.
Importantly, we implement retrieval merging strategy of SemanticGesticulator~\cite{zhang2024semanticgesture} to incorporate the output of our proposed retrieval algorithms into this training-based approach.
We perform this experiment by training it on all speakers and utilizing Discourse-based retrieval algorithm.
% To ablate the training-time RAG, we further modify the ReMoDiffuse architecture to use our gesture encoding and our retrieval algorithms, which is referred to as ``ReMoDiffuse+Our Retrieval''.
% 
% Analysis on this is given in the Ablative Analysis section~().
% 
\subsection{Runtime Information:}
Retrieval algorithms involve multiple ranking and filtering steps, each contributing to a certain computation time. 
Specifically, Discourse-based algorithm takes \textbf{0.03s} to run on 1 data sample.
LLM-based algorithm also includes an API call along with ranking steps and therefore, total time taken is increased to \textbf{1.52s} which includes \textbf{0.95s} for API call.
% Retreival for 1 sample: Discourse: \textbf{0.03s}; LLM-based: \textbf{1.52s} (incl. 0.95s for API call).
% 
RAG-driven inference for the diffusion model takes \textbf{26.93s} on NVIDIA RTX3090 for a batch size of 32.
% Diffusion Inference (batch size:32): \textbf{26.93s} on RTX3090.
% 
% {
%     \small
%     \bibliographystyle{ieeenat_fullname}
%     \bibliography{main}
% }

\end{document}